\documentclass[sigconf]{acmart}
\usepackage{amsmath,amsfonts}
\usepackage{algorithmic}
\usepackage{textcomp}
\usepackage{stfloats}
\usepackage{url}
\usepackage{verbatim}
\usepackage{hyperref}
\usepackage{epsfig}
\usepackage{amssymb}
\usepackage{multirow}
\usepackage{subfigure}
\usepackage{wrapfig}
\usepackage{booktabs}
\usepackage{pifont}
\usepackage{makecell}
\usepackage{balance}
\usepackage[lined,boxed,ruled, commentsnumbered]{algorithm2e}
\AtBeginDocument{%
  }

\copyrightyear{2024}
\acmYear{2024}
\setcopyright{acmlicensed}
\acmConference[KDD '24]{Proceedings of the 30th ACM SIGKDD Conference on Knowledge Discovery and Data Mining}{August 25--29, 2024}{Barcelona, Spain}
\acmBooktitle{Proceedings of the 30th ACM SIGKDD Conference on Knowledge Discovery and Data Mining (KDD '24), August 25--29, 2024, Barcelona, Spain}
\acmDOI{10.1145/3637528.3671681}
\acmISBN{979-8-4007-0490-1/24/08}

\begin{document}

\title{ITPNet: Towards Instantaneous Trajectory
Prediction for Autonomous Driving}

\author{Rongqing Li}
\affiliation{
  \institution{Beijing Institute of Technology}
  \city{Beijing}
  \country{China}
  }
\email{lirongqing99@gmail.com}

\author{Changsheng Li}
\authornote{Corresponding author}
\affiliation{
  \institution{Beijing Institute of Technology}
  \city{Beijing}
  \country{China}
  }
\email{lcs@bit.edu.cn}

\author{Yuhang Li}
\affiliation{
  \institution{Beijing Institute of Technology}
    \city{Beijing}
  \country{China}
  }
\email{596983629@qq.com}

\author{Hanjie Li}
\affiliation{
  \institution{Beijing Institute of Technology}
    \city{Beijing}
  \country{China}
  }
\email{lihanjieyouxiang@sina.com}

\author{Yi Chen}
\affiliation{
  \institution{Beijing Institute of Technology}
    \city{Beijing}
  \country{China}
  }
\email{2812232328@qq.com}

\author{Ye Yuan}
\affiliation{
  \institution{Beijing Institute of Technology}
    \city{Beijing}
  \country{China}
  }
\email{yuan-ye@bit.edu.cn}

\author{Guoren Wang}
\affiliation{
  \institution{Beijing Institute of Technology}
    \city{Beijing}
  \country{China}
  }
\email{wanggrbit@126.com}

\renewcommand{\shortauthors}{Rongqing Li et al.}

\begin{abstract}
  Trajectory prediction of moving traffic agents is crucial for the safety of autonomous vehicles, whereas previous approaches usually rely on sufficiently long-observed trajectory (e.g., 2 seconds) to predict the future trajectory of the agents. However, in many real-world scenarios, it is not realistic to collect adequate observed locations for moving agents, leading to the collapse of most prediction models. For instance, when a moving car suddenly appears and is very close to an autonomous vehicle because of the obstruction, it is quite necessary for the autonomous vehicle to quickly and accurately predict the future trajectories of the car with limited observed trajectory locations.  In light of this, we focus on investigating the task of instantaneous trajectory prediction, i.e., two observed locations are available during inference. To this end, we put forward a general and plug-and-play instantaneous trajectory prediction approach, called ITPNet. Specifically, we propose a backward forecasting mechanism to reversely predict the latent feature representations of unobserved historical trajectories of the agent based on its two observed locations and then leverage them as complementary information for future trajectory prediction. Meanwhile, due to the inevitable existence of noise and redundancy in the predicted latent feature representations, we further devise a Noise Redundancy Reduction Former (NRRFormer) module, which aims to filter out noise and redundancy from  unobserved trajectories and integrate the filtered features and observed features into a compact query representation for future trajectory predictions. In essence, ITPNet can be naturally compatible with existing trajectory prediction models, enabling them to gracefully handle the case of instantaneous trajectory prediction. Extensive experiments on the Argoverse and nuScenes datasets demonstrate ITPNet outperforms the baselines by a large margin and shows its efficacy with different trajectory prediction models.
\end{abstract}





\maketitle

\section{Introduction}
Predicting the future trajectories of dynamic traffic agents is a critical task for autonomous driving, which can be beneficial to the downstream planning module of  autonomous vehicles. 
In recent years, many trajectory prediction methods have been proposed in  computer vision and machine learning communities
\cite{wang2023prophnet, park2023leveraging,zhou2023query,zhu2023ipcc,chen2021human,gu2022stochastic,Xu_2022_CVPR,wang2022ltp,meng2022forecasting}. Among these methods, they usually need to collect sufficiently long observed trajectories (typically, 2 to 3 seconds) of an agent, in order to accurately predict its future trajectories.
 Recent advances have shown promising performance in trajectory prediction by learning from these adequate observations.

\begin{figure*}[t]
    \centering
\setcounter{subfigure}{0}
\subfigure[]{
		\begin{minipage}[t]{0.3\linewidth}
			\centering
			\includegraphics[width=1\linewidth]{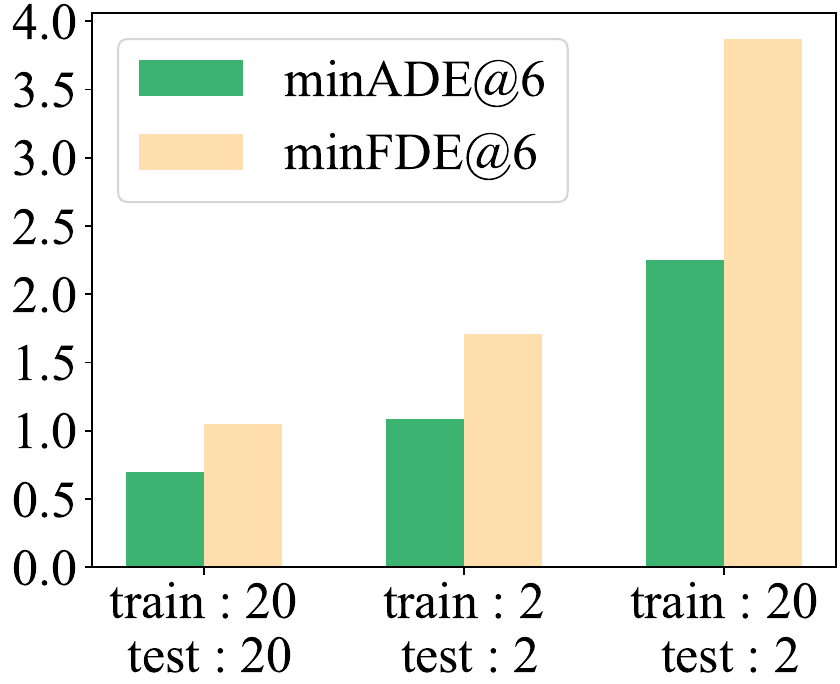}
		\end{minipage}
	}
    \subfigure[]{
		\begin{minipage}[t]{0.5\linewidth}
			\centering
			\includegraphics[width=1\linewidth]{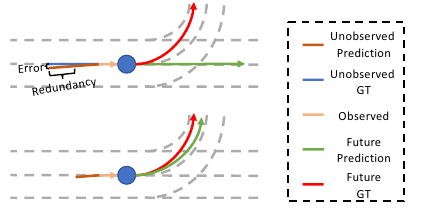}
		\end{minipage}
	}
 \vspace{-0.15in}
\caption{(a) Results of HiVT \cite{zhou2022hivt} in terms of minADE@6 and minFDE@6 on the validation set of Argoverse \cite{chang_argoverse_2019} with different observed locations as inputs during training and testing. The value in the horizontal axis denotes the number of observed locations. (b) Future predictions (shown in green) when utilizing different lengths of predicted unobserved trajectory locations.
The observed  trajectories are shown in orange, the predicted unobserved trajectories are shown in brown, the ground-truth unobserved trajectories are shown in blue, and the ground-truth future trajectories are shown in red.}
	\label{fig:intro}
 \vspace{-0.15in}
\end{figure*}

However,in real-world self-driving scenarios, it is often difficult to accurately predict trajectories due to the limited availability of observed locations. For instance, due to the obstruction,  a moving car might suddenly appear and be very close to the autonomous vehicle. At this  moment, the autonomous vehicle does not have enough time to collect adequate observed locations of the car to accurately predict the vehicle's future trajectories. Such a case will cause the collapse of the aforementioned prediction models due to the lack of information.
To verify this point,  we perform a typical trajectory prediction method, HiVT  \cite{zhou2022hivt}, with different  settings on the Argoverse dataset \cite{chang_argoverse_2019}. As shown in the left part of Figure \ref{fig:intro}(a), if we use 20 observed locations as the inputs of the prediction model during both training and test phases as in \cite{zhou2022hivt}, the prediction results are 0.698 and 1.053 in terms of minADE@6 and minFDE@6, respectively.
However, if we set only 2 observed locations as the inputs of the model during testing, the model will degrade sharply, no matter if the number of observed locations is 2 or 20 during the training phase. 
Thus, it is essential to study the trajectory prediction task, when observed locations are very limited. 

In light of this, we focus on studying the task of instantaneously predicting future trajectories of moving agents, under the assumption of only 2 trajectory locations available. Recently, The work in \cite{sun2022human} proposes a trajectory prediction method based on momentary observations. However, this method mainly focuses on the trajectory prediction of pedestrians, which has not been explored for other moving agents. In addition, the input of their model is the RGB image which usually contains  abundant context and semantic information. Thus, it is much easier for the model to predict future trajectories using RGB images,  compared to only several discrete trajectory locations. Moreover, \cite{monti2022many} design a knowledge distillation mechanism based on limited observed locations and achieves promising results. Since the method needs to pre-train a teacher model, and learns a student model distilling knowledge from the teacher model, which largely increases the computational complexities.

To this end, we propose a general and principled approach, called ITPNet, for instantaneous trajectory prediction by only two observed trajectory locations. Specifically, ITPNet aims to train a predictor to backwardly predict the latent feature representations of unobserved historical trajectory locations of the agent based on its two observed trajectory locations. The additional information contained in the predicted unobserved trajectory features assists observed trajectory features in better predicting future trajectories.
Nevertheless, we find that as we increase the number of backwardly predicted unobserved trajectory locations, the model's performance initially improves but subsequently deteriorates (This is verified in Table \ref{table:ablation} of Experiment section). We analyze two primary factors that impede the utilization of more unobserved trajectory features:
One is the noise brought by inaccurate prediction of the unobserved trajectory features. The other is a negative impact on the trajectory prediction due to  the intrinsic redundant information.
Let's consider a scenario where a vehicle  travels straightly for a while and then suddenly executes a turn. In such a case, a longer historical trajectory may erroneously boost the model's confidence in the vehicle continuing straight in the future, as depicted in the upper portion of Figure \ref{fig:intro}(b).
Conversely, a shorter unobserved historical trajectory with less redundancy tends to yield more accurate predictions because it maintains lower confidence in the vehicle's persistence in a straight trajectory and, instead, maintains higher confidence in the vehicle's persistence in a turning trajectory, as shown in the lower portion of Figure \ref{fig:intro}(b).
Thus, how to remove noisy and redundant information from the predicted features of the unobserved trajectories becomes the key to success in instantaneous trajectory prediction.

In view of this, we devise a Noise Redundancy Reduction Former (NRRFormer) module and integrate it into our framework. NRRFormer can filter out noise and redundancy from a sequence of predicted unobserved latent features, and effectively fuse the filtered unobserved latent features with the observed features by a compact query embedding to acquire the most useful information for future trajectory prediction. 
It is worth noting that our ITPNet is actually plug-and-play, and is compatible with existing trajectory prediction models, making them the kinds that can gracefully deal with the instantaneous trajectory prediction problem. 

Our main contributions are summarized as:
1) We design a backward forecasting mechanism to reconstruct unobserved historical trajectory information for instantaneous trajectory prediction, mitigating the issue of lack of information due to only two observed locations. 
2) We devise a Noise Redundancy Reduction Former
(NRRFormer), which can remove noise and redundancy among the predicted unobserved features to further improve the prediction performance. 
3) We perform extensive experiments on two widely used benchmark datasets, and demonstrate ITPNet can outperform baselines in a large margin. Moreover, we show the efficacy of ITPNet, combined with different trajectory prediction models.

\section{Related Works}
\subsection{Trajectory Prediction with Sufficient Observation}

In recent years, many trajectory prediction approaches have been proposed \cite{girgis2021latent,gilles2022thomas,makansi2021you,casas2020spagnn,sun2022m2i,cheng2023forecast,bae2023eigentrajectory,bae2023graphtern,choi2023r}. In the early stage of trajectory prediction, studies such as \cite{alahi2016social, gupta2018social} usually rely solely on observation points and adopt simple social pooling methods to capture interactions between agents. To capture the map information, including occupancy or semantic information, \cite{bansal2018chauffeurnet,Phan-Minh_2020_CVPR,mohamed2020social} propose to use Convolutional Neural Networks (CNN) to encode map images. In addition, \cite{gao2020vectornet,liang2020learning} incorporate the information of lanes and traffic lights on the map in the form of vectors.
Recently,  numerous methods have been proposed to fully exploit the interaction information between nearby agents, including implicit modeling by graph neural networks \cite{casas2020spagnn,li2019grip++,salzmann2020trajectron++,casas2020implicit} and attention mechanisms \cite{nayakanti2022wayformer,liu2021multimodal,ngiam2022scene,li2020end}, and explicit modeling \cite{sun2022m2i}.
To handle the uncertainty of road agents, researchers propose to generate multi-modal trajectories using various approaches, including GAN-based methods \cite{kosaraju2019social,sadeghian2019sophie,gupta2018social}, VAE-based methods \cite{lee2017desire,lee2022muse}, flow-based methods \cite{zhang2022human,liang2022stglow}, and diffusion models \cite{gu2022stochastic,mao2023leapfrog,jiang2023motiondiffuser}. Among them, one typical approach is to establish a mapping between future trajectories and latent variables, producing multiple plausible trajectories by sampling the latent variable. 
In addition, goal-based methods have become popular recently \cite{zhao_tnt_2020,gu_densetnt_2021,zeng2021lanercnn,wang2022ganet,mangalam_goals_2021,Aydemir2023ICCV}, which first generates multi-modal goals by sampling \cite{zhao_tnt_2020} or learning \cite{wang2022ganet}, and then predict future trajectories conditioned on the goals. 

Although these methods have shown promising performance in trajectory prediction, they usually learn depending on sufficiently long-observed locations. As aforementioned, these methods degrade severely or even collapse when the number of observed locations is limited. 
Different from these works,  we attempt to address the task of instantaneously predicting the future trajectories of moving agents, under the condition that only two trajectory locations are observable.

\subsection{Trajectory Prediction with Instantaneous Observation}

Predicting the future trajectories of a moving agent by its limited observed locations remains a challenging problem. Recently, \cite{sun2022human} proposes an approach to integrate the velocity of agents, social and scene contexts information, and designs a momentary observation feature Extractor (MOE) for pedestrian trajectory prediction. 
The input of MOE contains image frames from videos which usually contain abundant semantic information. Thus, it is much easier to predict future trajectories using image frames than that using several discrete trajectory points. 
Moreover, since this method is mainly designed for predicting trajectories for pedestrians, what is the performance on other moving agents, e.g., cars, is worth to be further verified. 
 \cite{monti2022many} proposes a knowledge distillation approach using few observations as input, with the goal of lowering the influence of noise introduced by the machine perception side (i.e., incorrect detection and tracking).
As we know, knowledge distillation-based approaches generally need to pre-train a teacher model, and then distill knowledge from the teacher model to help the student model learn, which makes this kind of method computationally expensive.
\cite{bae2024singulartrajectory} designed a unified framework for tackling multiple tasks including instantaneous pedestrian trajectory prediction, where they constructed a universal singular space to share the information between long and short trajectory points. Differently, our focus lies on the task of instantaneous trajectory prediction, achieved by predicting unobserved historical trajectories, without using any information of long trajectory points.

\section{Proposed Method}

\label{pm}
\begin{figure*}
\setlength{\belowcaptionskip}{-0.4cm} 
  \begin{center}
  \includegraphics[width=0.95\linewidth]{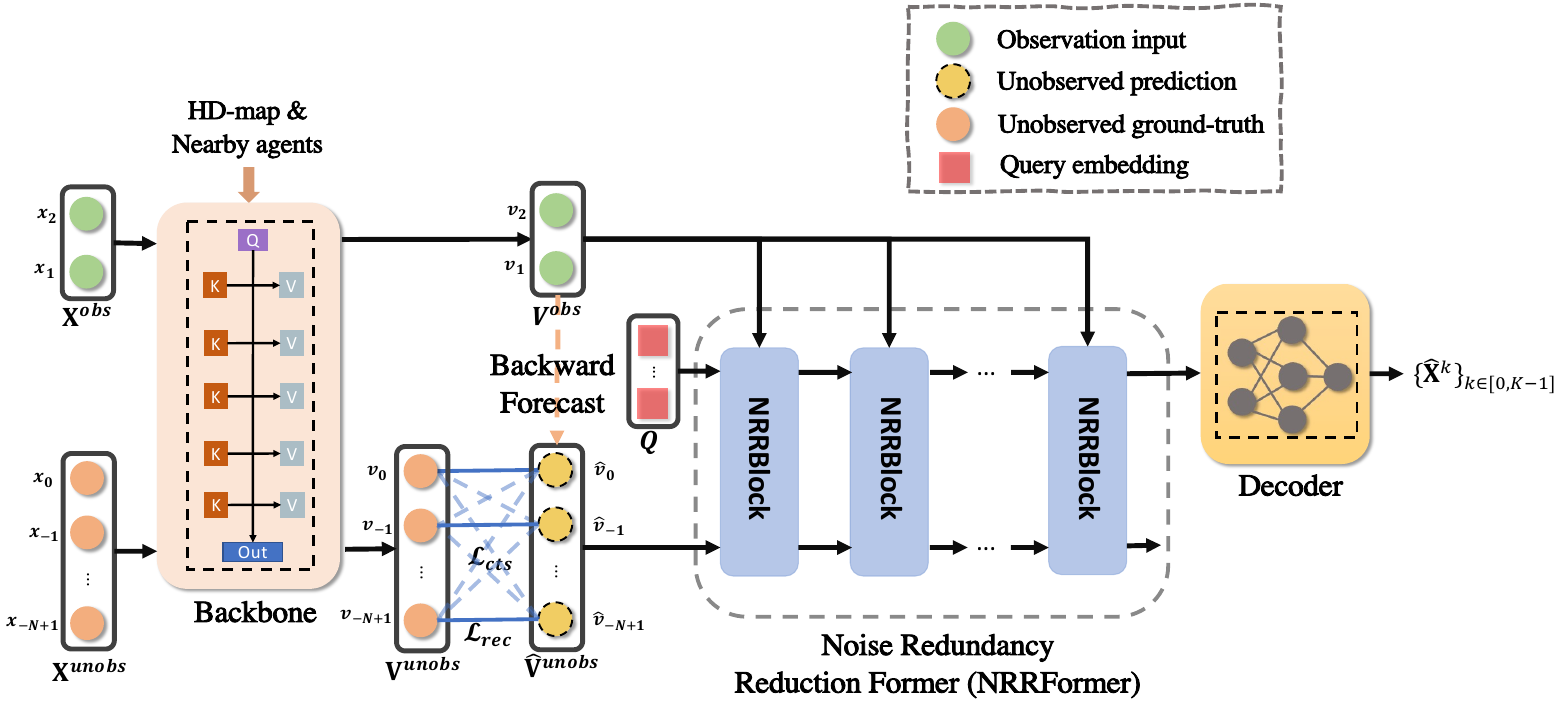}
  \caption{Overview of our ITPNet framework. ITPNet mainly consists of two modules: 1) We propose a backward forecasting mechanism that attempts to reconstruct the latent feature representations $\mathbf{V}^{unobs}$ of previous unobserved trajectory locations $\mathbf{X}^{unobs}$  by the two  observed trajectories locations $\mathbf{X}^{obs}$. 2) We devise a Noise Redundancy Reduction Former to filter out noise and redundancy in the predicted latent feature representations $\hat{\mathbf{V}}^{unobs}$, and both the resulting filtered features and the observation features $\mathbf{V}^{obs}$ are integrated into a compact query embedding $\mathbf{Q}$. Finally, the query embedding is sent to the decoder to instantaneously predict future trajectories $\{\hat{\mathbf{X}}^k\}.$
  } 

  \label{fig:framework}
  \end{center}
\end{figure*}

\subsection{Problem Definition}

We denote a sequence of observed trajectory locations for a target vehicle as $\mathbf{X}^{obs}\!=\!\{x_{1}, x_{2}, ..., x_{T}\}$,
where $x_{i} \in \mathbb{R}^2$ is the $i$-th  location of the agent, and $T$ represents the number of observed historical locations. We set $T=2$ for the most extreme case, aligning with the setting of MOE \cite{sun2022human}, which is the first work of instantaneous trajectory prediction.
Moreover, we also denote the sequence of previous unobserved trajectory locations of the agent as $\mathbf{X}^{unobs}\!=\!\{x_{-N+1}, x_{-N+2},\cdots, x_{0}\}$, where $N$ is the total number of unobserved trajectory locations.
The ground-truth future trajectories are denoted as $\mathbf{X}^{gt}\!=\!\{x_{3}, x_{4}, ..., x_{2+M}\}$, where $M$ is the length of ground-truth future trajectory. 
Our goal is to predict $K$ plausible trajectories $\{{\widehat{\mathbf{X}}}^k\}_{k \in [0, K-1]}\!=\!\{(\hat{x}_{3}^k, \hat{x}_{4}^k, ..., \hat{x}_{2+M}^k)\}_{k \in [0, K-1]}$, as in multi-model trajectory prediction methods \cite{gupta2018social,kosaraju2019social,lee2022muse}. In contrast to previous methods utilizing sufficient observed trajectory locations (typically, 20 observed trajecotory locations on the Argoverse dataset \cite{chang_argoverse_2019}), we attempt to leverage merely $T=2$ observed locations $\mathbf{X}^{obs}$ for instantaneous trajectory prediction.  It is noteworthy that, in principle, our proposed method is capable of accepting observed trajectory locations of arbitrary length $T$ as input.


\subsection{Overall Framework of ITPNet}
Figure \ref{fig:framework} illustrates an overview of our proposed framework.
We first feed the instantaneous observed trajectory locations $\mathbf{X}^{obs}$ into a backbone (e.g., HiVT  \cite{zhou2022hivt}) to obtain the latent feature representations $\mathbf{V}^{obs} = \{v_{1}, v_{2}\}$. Based on this representation $\mathbf{V}^{obs}$, we then attempt to backwardly predict the latent feature representations $\widehat{\mathbf{V}}^{unobs} = \{\hat{v}_{-N+1},$ $\hat{v}_{-N+2}, ..., \hat{v}_{0}\}$ of unobserved historical trajectories  $\mathbf{X}^{unobs}$. Considering that the predicted unobserved feature representations $\widehat{\mathbf{V}}^{unobs}$ inevitably contain redundant and noisy information as mentioned above, we design a Noise Redundancy Reduction Former (NRRFormer) module to filter out this information from a predicted feature sequence.   Subsequently, the filtered features are combined with the observed  features to generate a compact query embedding $\mathbf{Q}$. The query embedding $\mathbf{Q}$ is then sent to the decoder for future trajectory predictions.
Since the backbone in our framework is arbitrary, our method is plug-and-play, and is compatible with existing trajectory prediction models, enabling them to gracefully adapt to the scenario of instantaneous trajectory prediction.
In the following section, we will delve into a detailed introduction of backward forecasting and the NRRFormer.


\subsection{Backward Forecasting}
When given only 2 observed locations $\mathbf{X}^{obs}$, one major issue we face is the lack of information, making existing trajectory prediction approaches degraded sharply. To alleviate this problem, we propose to backwardly predict the latent feature representations of previous unobserved trajectory locations, and then leverage them as additional information for future trajectory prediction. 

First, we can obtain the latent feature representations $\mathbf{V}^{obs}$ of the observed locations $\mathbf{X}^{obs}$ via the backbone $\Phi$:
\begin{equation}
    \mathbf{V}^{obs}=\{v_1,v_2\} =\Phi(\mathbf{X}^{obs};\phi),
\end{equation}
where $v_{i} \in \mathbb{R}^d$ is the latent feature representation of the $i$-th location of the agent, and $d$ is the dimension of the feature. The backbone $\Phi$ is parameterized by $\phi$, and can be an arbitrary trajectory prediction model, e.g., HiVT \cite{zhou2022hivt} and LaneGCN \cite{liang2020learning} used in this paper.

After that, we attempt to backwardly predict the latent feature representations $\widehat{\mathbf{V}}^{unobs}$ on the basis of $\mathbf{V}^{obs}$, addressing the issue of the lack of information. To this end, we introduce two self-supervised tasks: the first one is the reconstruction of the latent feature representations, and the loss function is designed as:
\begin{equation}
   \mathcal{L}_{rec} = \mathcal{J} (\mathbf{V}^{unobs};\widehat{\mathbf{V}}^{unobs}),   
\end{equation}
where $\mathbf{V}^{unobs} = \Phi(\mathbf{X}^{unobs};\phi)$ is the ground-truth latent feature representations of previous unobserved trajectory locations, and can be taken as a self-supervised signal, and $\mathcal{J}$ is a function to measure the distance between $\mathbf{V}^{unobs}$ and $\widehat{\mathbf{V}}^{unobs}$.  
$\widehat{\mathbf{V}}^{unobs}$ are the predicted features, obtained by:
\begin{equation}
\widehat{\mathbf{V}}^{unobs} =  \Psi(\mathbf{V}^{obs};\psi),
\end{equation}
where $\Psi$ is a network parameterized by $\psi$. 
In this paper, we make use of a LSTM \cite{hochreiter1997long} to predict the $\widehat{\mathbf{V}}^{unobs}$ on the basis of $\mathbf{V}^{obs}$,


\begin{align}
    \hat{v}^{unobs}_{0} &= \Psi(\mathbf{Mean}(\mathbf{V}^{obs}), \psi), \notag \\
    \hat{v}^{unobs}_{i-1} &= \Psi(\hat{v}^{unobs}_{i};\psi), i=0,...,-N+2,
\end{align}
where $\hat{v}^{unobs}_i$ is the $i^{th}$ predicted unobserved latent feature representations of $\widehat{\mathbf{V}}^{unobs}$, and 
$\textbf{Mean}(\cdot)$ represents averaging the trajectory features along the length dimension.
In order to reconstruct the latent feature representations, we use the smooth $L_1$ loss \cite{girshick2015fast} to optimize the $\mathcal{L}_{rec}$ as:
\begin{equation}
   \label{l_rec}
   \mathcal{L}_{rec} = \sum_{i=-N+1}^{0}\delta(v_{i}^{unobs} - \hat{v}_{i}^{unobs}),
\end{equation}
where $\delta$ is defined as:
\begin{equation}
\label{equ:delta}
\delta(v) =\left\{
\begin{array}{rcl}
0.5v^2 & & {if \ ||v|| < 1}\\
||v||-0.5 & & {otherwise,}\\
\end{array} \right.
\end{equation}
where $||v||$ denotes the $l_1$ norm of $v$.

To further enhance the representation ability of the unobserved latent feature representations, we devise another self-supervised task.
Specifically, we regard the feature pair $\{v_{i}^{unobs}, \hat{v}_{i}^{unobs}\}$ as the positive sample pair, $i=-N+1,\cdots, 0$, and take $\{v_i^{unobs}, \hat{v}_j^{unobs}\}$ as the negative sample pair, $i\neq j$. 
After that, we present another self-supervised loss:
\begin{equation}
\label{cts}
\mathcal{L}_{cts} = \sum_{i=-N\!+\!1}^0 \sum_{j\neq i} \max(0, \delta(v_i^{unobs} \!- \!\hat{v}_i^{unobs}) \!- \!\delta(v_i^{unobs} \!- \!\hat{v}_{j}^{unobs}) \!+ \!\Delta),
\end{equation}
where $\Delta$ is a margin.  It is worth noting that the first loss $\mathcal{L}_{rec}$ in (\ref{l_rec}) targets at reconstructing the latent feature representation $v_i$ as accurately as possible, while the second loss $\mathcal{L}_{cts}$ in 
(\ref{cts}) aims to minimize the discrepancy between the predicted unobserved feature representations and the corresponding ground-truth feature representations at each timestep, while it enlarges a margin $\Delta$ between the predicted unobserved and non-corresponding ground-truth feature representations. This can further assist in better reconstructing unobserved trajectory locations.

\subsection{Noise Redundancy Reduction Former}
Our Noise Redundancy Reduction Former (NRRFormer) module that is parameterized by $\Theta$ contains $L$ Noise Redundancy Reduction Blocks (NRRBlocks). Each NRRBlock attempts to filter out noise and redundancy in the predicted latent feature representations $\hat{\mathbf{V}}_l^{unobs}$, and integrate the resulting filtered feature representations and observed feature representations $\mathbf{V}^{obs}$ into a query embedding $\mathbf{Q}_{l+1}, l=0,1,\cdots,L-1$.

\begin{figure}
  \begin{center}
  \vspace{-2mm}
    \includegraphics[width=0.7\linewidth]{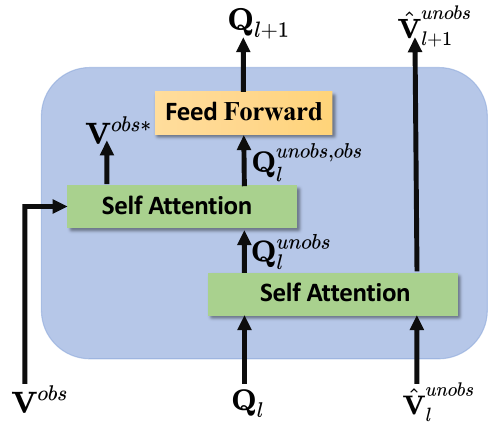}
  \end{center}
  \vspace{-5mm}
  \caption{Structure of Noise Redundancy Reduction Block.} 
    \label{fig:NRRBlock}
\end{figure}

As shown in the Figure \ref{fig:NRRBlock}, The $l^{th}$ layer of NRRBlock takes as input the query embedding $\mathbf{Q}_l$ and the unobserved feature representations  $\hat{\mathbf{V}}_l^{unobs}$ through a self-attention mechanism:
\begin{equation}
    \mathbf{Q}^{unobs}_l, \hat{\mathbf{V}}^{unobs}_{l+1} = \textbf{SelfAtt}(\mathbf{Q}_l || \hat{\mathbf{V}}^{unobs}_{l};\theta_{l,1}),
\end{equation}
where $||$ denotes the concatenation operation, the self-attention module is parameterized by $\theta_{l,1}$. $\mathbf{Q}_0$ is a random initialized tensor, $\hat{\mathbf{V}}_{0}^{unobs} = \hat{\mathbf{V}}^{unobs}$, and the $\mathbf{Q}^{unobs}_l$ represents the output query embedding. It is worth noting that the length of the query, denoted as $C$, is smaller than the length of $\hat{\mathbf{V}}^{unobs}_{l}$, denoted as $N$, so that information in $\hat{\mathbf{V}}^{unobs}_{l}$ is forced to condense and collate into the compact query embedding $\mathbf{Q}^{unobs}_l$, thereby filtering out redundancy and noise to extract the meaningful information. After that, we utilize another self-attention module to integrate information of $\mathbf{V}^{obs}$ into the query embedding:
\begin{equation}
    \label{selfatt2}
    \mathbf{Q}^{unobs,obs}_l, \mathbf{V}^{obs*} = \textbf{SelfAtt}(\mathbf{Q}^{unobs}_l || \mathbf{V}^{obs};\theta_{l,2}),
\end{equation}
where the self-attention module is parameterized by $\theta_{l,2}$, $\mathbf{Q}^{unobs,obs}_l$ represents the query embedding after integrating both the filtered unobserved trajectory features and the observed trajectory features. Through this self-attention operation, the information of $V^{obs}$ can be effectively distilled into Q, while enabling it to fuse with $V^{unobs}$, thereby facilitating the exchange of complementary information between them. Note that we assume the observed trajectory features $\mathbf{V}^{obs}$ do not contain noise or redundancy, because the features are obtained by encoding $\mathbf{X}^{obs}$. Therefore, the Equation (\ref{selfatt2}) only integrates the information of $\mathbf{V}^{obs}$ into the query $\mathbf{Q}$ through self-attention, but not input the $\mathbf{V}^{obs*}$ into the next NRRBlock.
At the end of the NRRBlock, we employ a feed forward layer to produce the query representation for the next layer,
\begin{equation}
    \mathbf{Q}_{l+1} = \textbf{FeedForward}(\mathbf{Q}_l^{unobs,obs};\theta_{l,3}),
\end{equation}
where the feed forward layer is parameterized by $\theta_{l,3}$.
We utilize $L$ NRRBlocks to denoise and reduce redundancy in the unobserved trajectory features while effectively fusing the observed trajectory features. Finally we utilize $\mathbf{Q}_L$ 
for future trajectory prediction:
\begin{equation}
\label{output}
    \{\widehat{\mathbf{X}}^k\}_{k\in [0,K\!-\!1]} \!=\! \Omega(\mathbf{Q}^L;\omega), 
\end{equation}
where $\Omega$ represents the decoder module parameterized by $\omega$.
The decoder module can be the same structure as in previous trajectory prediction models \cite{zhou2022hivt,liang2020learning}, enabling our method to be generalizable. 

\subsection{Optimization and Inference}
We adopt the commonly used winner-takes-all strategy \cite{zhao_tnt_2020} on the obtained $K$ multi-modal trajectories $\{\widehat{\mathbf{X}}^k\}_{k\in [0,K-1]}$,
which regresses the trajectory closest to the ground truth, denoted as $\mathcal{L}_{reg}$. In order to help the downstream planner make better decisions, a classification loss $\mathcal{L}_{cls}$ is also adopted to score each trajectory. Here, we adopt the same $\mathcal{L}_{reg}$ and $\mathcal{L}_{cls}$ as those in the corresponding backbones (see Appendix \ref{app:detail_loss} for details of $\mathcal{L}_{reg}$ and $\mathcal{L}_{cls}$).  Finally, the total loss function can be expressed as:
\begin{equation}
    \mathcal{L} = \mathcal{L}_{reg} + \mathcal{L}_{cls} + \alpha\mathcal{L}_{rec} + \beta\mathcal{L}_{cts},
\end{equation}
where $\alpha$ and $\beta$ are three trade-off hyper-parameters. We provide the pseudo-code of our training procedure in the Algorithm \ref{alg:train}.

\begin{algorithm}
  \caption{Training Procedure of ITPNet}
  
  \SetKw{Initialization}{Initialization}

    \KwIn{input trajectory $\mathbf{X}=\{\mathbf{X}^{obs}, \mathbf{X}^{unobs}\}$, ground-truth trajectory $\mathbf{X}^{gt}$, query embedding $\mathbf{Q}$, layers $L$ of NRRFormer, trade-off hyper-parameters: $\alpha$, and $\beta$.}
    
    \KwOut{Network parameters: $\phi$, $\psi$, $\{\theta_{l,1}, \theta_{l,2}, \theta_{l,3}\}_{l=1}^L$, and $\omega$.}

    {\textbf{Initialize:} Randomly initialize  $\phi$, $\psi$, $\{\theta_{l,1}, \theta_{l,2}, \theta_{l,3}\}_{l=1}^L$, $\omega$, and $\mathbf{Q}$.} \\
    \While{not converges}{
        Compute latent feature representations $\mathbf{V}^{obs} = \Phi(\mathbf{X}^{obs};\phi)$ and $\mathbf{V}^{unobs} = \Phi(\mathbf{X}^{unobs};\phi)$; \\
        Backward forecast $\widehat{\mathbf{V}}^{unobs}$ by $\widehat{\mathbf{V}}^{unobs}=\Psi(\mathbf{V}^{obs};\psi)$; \\ 
        Compute $\mathcal{L}_{rec},\mathcal{L}_{cts}$ by Eq. (\ref{l_rec}) and (\ref{cts}), respectively; \\
        Employ NRRFormer to filter out redundancy and noise in predicted unobserved latent feature representations and integrate the resulting filtered feature representations and observed feature representations into $Q$, by \\
        $\hat{\mathbf{V}}_0^{unobs} = \hat{\mathbf{V}}^{unobs}$; \\
        \For{$l=0...L-1$}{
            $\mathbf{Q}^{unobs}_l, \hat{\mathbf{V}}^{unobs}_{l+1} = \textbf{SelfAtt}(\mathbf{Q}_l || \hat{\mathbf{V}}^{unobs}_{l};\theta_{l,1})$ ;\\
            $\mathbf{Q}^{unobs,obs}_l, \mathbf{V}^{obs*} = \textbf{SelfAtt}(\mathbf{Q}^{unobs}_l || \mathbf{V}^{obs};\theta_{l,2})$; \\
            $\mathbf{Q}_{l+1} = \textbf{FeedForward}(\mathbf{Q}_l^{unobs,obs};\theta_{l,3})$;
        }
        Predict trajectory $\{\widehat{\mathbf{X}}^k\}_{k \in [0, K-1]} = \Omega(\mathbf{Q}_{L};\omega)$; \\
        Compute $\mathcal{L}_{reg},\mathcal{L}_{cls}$ through $\{\widehat{\mathbf{X}}^k\}_{k \in [0, K-1]}$;\\ 
        
        Calculate the total loss $\mathcal{L}$ by
        $\mathcal{L}=  \mathcal{L}_{reg} + \mathcal{L}_{cls} + \alpha \mathcal{L}_{rec}$ + $\beta \mathcal{L}_{cts}$;\\
        Update model parameters $\phi$, $\psi$, $\{\theta_{l,1}, \theta_{l,2}, \theta_{l,3}\}_{l=0}^{L-1}$, $\omega$ and query embedding $\mathbf{Q}$ by minimizing $\mathcal{L}$.
    }
    \label{alg:train}
\end{algorithm}

For inference, when only 2 observed trajectory locations of a target vehicle are collected, we first extract the latent feature representations based on the backbone $\Phi$,  and then apply our backward forecasting mechanism to predict the latent feature representations of previous $N$ unobserved locations of the target agent by the networks $\Psi$. After that, the NRRFormer $\Theta=\{\theta_{l,1},\theta_{l,2},\theta_{l,3}\}_{l=1}^L$ filters out the noise and redundancy in the unobserved latent feature representations and integrates the filtered features and observed latent feature representations into query embedding. Finally, the query embedding are fed into the decoder network $\Omega$ for instantaneous trajectory prediction.

\section{Experiments}

\begin{table*}[h]
\renewcommand\arraystretch{1.1}
 \setlength\tabcolsep{1.6pt}
\centering
\caption{minADE@$K$, minFDE@$K$, and MR@$K$ of different methods on Argoverse and nuScenes, respectively.}
\vspace{-0.1in}
\begin{tabular}{c|c|ccc|ccc}
\toprule[1.3pt]
\multirow{2}{*}{Dataset}& \multirow{2}{*}{Methods} & \multicolumn{3}{c|}{K=1} & \multicolumn{3}{c}{K=6} \\
& & minADE & minFDE & minMR & minADE & minFDE & minMR\\ \hline

\multirow{8}{*}{Argoverse}
& HiVT \citep{zhou2022hivt} & 4.158 & 8.368 & 0.846 & 1.085 & 1.712 & 0.249 \\
& MOE+HiVT \citep{sun2022human} & 3.312 & 6.840 & 0.794 & 0.939 & 1.413 & 0.177 \\ 
& Distill+HiVT \citep{monti2022many} & 3.251 & 6.638 & 0.771 & 0.968 & 1.502 & 0.185 \\ 
& \textbf{ITPNet+HiVT} & \textbf{2.631} & \textbf{5.703} & \textbf{0.757} & \textbf{0.819} & \textbf{1.218} & \textbf{0.141} \\ 
\cline{2-8}
& LaneGCN \citep{liang2020learning} & 4.204 & 8.647 & 0.861 & 1.126 & 1.821 & 0.278 \\
& MOE+LaneGCN \citep{sun2022human} & 3.958 & 8.264 & 0.842 & 1.089 & 1.734 & 0.265 \\ 
& Distill+LaneGCN \citep{monti2022many} & 3.768 & 7.926 & 0.817 & 1.077 & 1.687 & 0.252 \\ 
& \textbf{ITPNet+LaneGCN} &  \textbf{2.922}& \textbf{5.627} & \textbf{0.765} & \textbf{0.894} & \textbf{1.425} & \textbf{0.173} \\

\hline \hline
\multirow{8}{*}{nuScenes}
& HiVT \citep{zhou2022hivt} & 6.564 & 13.745 & 0.914 & 1.772 & 2.836 & 0.505 \\ 
& MOE+HiVT \citep{sun2022human} & 5.705 & 12.619 & 0.913 & 1.712 & 2.813 & 0.494 \\ 
& Distill+HiVT \citep{monti2022many} & 5.950 & 12.606 & 0.911 & 1.759 & 2.861 & 0.483 \\
& \textbf{ITPNet+HiVT} & \textbf{5.514} & \textbf{12.584} & \textbf{0.909} & \textbf{1.503} & \textbf{2.628} & \textbf{0.483} \\ 
\cline{2-8}
& LaneGCN \citep{liang2020learning} & 6.125 & 14.300 & 0.935 & 1.878 & 3.497 & 0.630 \\
& MOE+LaneGCN \citep{sun2022human} & 6.071 & 13.994 & 0.931 & 1.778 & 3.372 & 0.613 \\ 
& Distill+LaneGCN \citep{monti2022many} & 5.968 & 13.807 & 0.926 & 1.737 & 3.278 & 0.604 \\
& \textbf{ITPNet+LaneGCN} &\textbf{5.739} & \textbf{13.555} & \textbf{0.919} & \textbf{1.679} & \textbf{3.146} & \textbf{0.580} \\

\bottomrule[1.3pt]
\end{tabular}
\label{table:sota_comp}
\end{table*}

\subsection{Datasets}
We evaluate our method for the instantaneous trajectory prediction tasks on two widely used benchmark datasets, Argoverse \citep{chang_argoverse_2019} and NuScene \citep{caesar2020nuScenes}.

\noindent\textbf{Argoverse Datasets:}
This dataset contains a total of 324,557 scenes, which are split into 205,492 training scenes, 39,472 validation scenes, and 78,143 testing scenes. 
The observation duration for both the training and validation sets is 5 seconds with a sampling frequency of 10Hz. In contrast to previous approaches taking the first 2 seconds (i.e., 20 locations) as the observed trajectory locations and the last 3 seconds as the future ground-truth trajectory, we only utilize $T=2$ observed trajectory locations, and predict the future trajectory of the last 3 seconds in our experiments.

\noindent\textbf{NuScene Datasets}
The dataset consists of 32,186 training, 8,560 validation, and 9,041 test samples.
Each sample is a sequence of x-y coordinates with a duration of 8 seconds and a sample frequency of 2Hz. Previous approaches usually take the first 2 seconds (i.e., 5 locations) as the observed trajectory locations
and the last 6 seconds as the future ground-truth trajectory. However, we leverage only $T=2$ observed trajectory locations to predict the future trajectory of the last 6 seconds in the experiments. 


\subsection{Implementation Details}
\label{chapter:impl_details}
We perform the experiments using two different backbone models, HiVT \citep{zhou2022hivt} and LaneGCN \citep{liang2020learning}. 
Specifically, we utilize the temporal encoder in HiVT and the ActorNet in LaneGCN to extract the latent feature representations, respectively.
We set the feature dimensions $d$ to 64 and 128 when using HiVT and LaneGCN as the backbone, respectively.  The hidden size of the LSTM for predicting unobserved latent feature representations is set to $d$. The NRRFormer consists of three NRRBlocks. In our experiments, the predicted unobserved length $N$ is set to 10 for the Argoverse dataset and 4 for the nuScenes dataset. We set the query embedding length to $C=4$ for the Argoverse dataset and $C=2$ for the nuScenes dataset. In addition, we set the trade-off hyper-parameters $\alpha$ and $\beta$ to 0.1 and 0.1.


\subsection{Baselines and Evaluation Metrics}
We first compare with two most related works: MOE \citep{sun2022human} and Distill \citep{monti2022many}. 
Since we use HiVT \citep{zhou2022hivt} and LaneGCN \citep{liang2020learning} as our backbone, respectively, we also compare our method with them.
When using HiVT as the backbone, we denote our method as ITPNet+HiVT.
When using LaneGCN as the backbone, we denote our method as ITPNet+LaneGCN. 

To evaluate these methods, we employ three popular evaluation metrics \citep{zhao_tnt_2020, gu_densetnt_2021,wang2022ganet},  minADE@$K$, minFDE@$K$, and minMR@$K$, where $K$ represents the number of the generated trajectories. we set $K$ to 1 and 6 in our experiments.

\begin{table*}[t]
\centering
\caption{Ablation study of our method for $\mathcal{L}_{rank}$, $\mathcal{L}_{rec}$ and $\mathcal{L}_{cts}$ on the Argoverse dataset. }
\vspace{-0.1in}
\begin{tabular}{c|c|c|ccc|ccc}
\toprule[1.3pt]
\multirow{2}{*}{$\mathcal{L}_{rec}$}  & \multirow{2}{*}{$\mathcal{L}_{cts}$} & \multirow{2}{*}{NRRFormer} & \multicolumn{3}{c|}{K=1} & \multicolumn{3}{c}{K=6} \\
& & & minADE & minFDE & minMR & minADE & minFDE & minMR\\ \hline
& &  &  4.158 & 8.368 & 0.846 & 1.085 & 1.712 & 0.249 \\
\checkmark & &  & 2.646 & 5.790 & 0.763 & 0.841 & 1.285 & 0.154 \\ 
\checkmark & \checkmark &  & \textbf{2.615} & 5.733 & 0.761 & 0.832 & 1.262 & 0.149 \\ 

\checkmark & \checkmark & \checkmark  & 2.631 & \textbf{5.703} & \textbf{0.757} & \textbf{0.819} & \textbf{1.218} & \textbf{0.141} \\ 
\bottomrule[1.3pt]
\end{tabular}
\label{table:ablation1}
\end{table*}

\subsection{Results and Analysis}
\noindent\textbf{Performance on Instantaneous Trajectory Prediction}.
To demonstrate the effectiveness of our method for instantaneous trajectory prediction, we compare our method with the state-of-the-art baselines. The results listed in Table \ref{table:sota_comp}, shows that ITPNet+LaneGCN and ITPNet+HiVT significantly outperforms LaneGCN and HiVT, respectively. 
This illustrates that current state-of-the-art trajectory prediction approaches struggle to effectively handle cases involving instantaneous observed trajectory inputs. However, when plugging our framework into these two backbone models, respectively, the performance is significantly improved. This shows our method is effective for instantaneous trajectory prediction, and is compatible with different trajectory prediction models. Moreover, our methods achieve better performance than MOE and Distill, which indicates the effectiveness of our methods once more. 


\begin{table*}[ht]
\renewcommand\arraystretch{1.0}

\centering
\caption{Analysis of backward forecasting with different $N$ and effectiveness of NRRFormer on Argoverse.}
\vspace{-0.1in}
\begin{tabular}{c|c|ccc|c|ccc}
\toprule[1.3pt]
\multirow{2}{*}{N} & \multirow{2}{*}{\makecell[c]{NRR\\Former}} & \multicolumn{3}{c|}{K=6} & \multirow{2}{*}{\makecell[c]{NRR\\Former}} & \multicolumn{3}{c}{K=6} \\
 & & minADE & minFDE & minMR & &  minADE & minFDE & minMR\\ \hline
  0 & \ding{55}  & 1.068 & 1.678 & 0.241 & - & - & - & - \\ 
  1 & \ding{55}   & 0.969 & 1.494 & 0.193 & \checkmark & 0.964& 1.498& 0.194\\ 
  2 & \ding{55}   & 0.872 & 1.329 & 0.160 & \checkmark & 0.868& 1.323& 0.158\\
  3 & \ding{55}  & 0.832 & 1.262 & 0.149 & \checkmark & 0.828& 1.254& 0.147\\ 
  4 & \ding{55}  & 0.845 & 1.291 & 0.154 & \checkmark & 0.824& 1.240& 0.146\\ 
  5 & \ding{55}  & 0.859 & 1.312 & 0.156 & \checkmark & 0.822& 1.232& 0.145\\ 
  6 & \ding{55}  & 0.867 & 1.302 & 0.161 & \checkmark & 0.823& 1.231& 0.145\\ 
  7 & \ding{55}  & 0.881 & 1.375 & 0.173 & \checkmark & 0.820& 1.222&0.143 \\ 
  8 & \ding{55}  & 0.903 & 1.410 & 0.181 & \checkmark & 0.821& 1.222&0.142 \\ 
  9 & \ding{55}  & 0.933 & 1.453 & 0.187 & \checkmark & 0.819& 1.220& 0.142\\ 
  10 &\ding{55}  & 0.967 & 1.522 & 0.196 & \checkmark & 0.819 & 1.218 & 0.141 \\
\bottomrule[1.3pt]
\end{tabular}
\label{table:ablation}
\end{table*}

\begin{table*}[ht]
\renewcommand\arraystretch{1.1}
 \setlength\tabcolsep{1.6pt}
\centering
\caption{Comparison with baselines on Argoverse dataset using different $T$}
\vspace{-0.1in}
\begin{tabular}{c|c|c|ccc|ccc}
\toprule[1.3pt]
\multirow{2}{*}{Dataset} & \multirow{2}{*}{T} & \multirow{2}{*}{Method} & \multicolumn{3}{c|}{K=1} & \multicolumn{3}{c}{K=6} \\
& &  &  minADE & minFDE & minMR & minADE & minFDE & minMR\\ \hline

\multirow{9}{*}{Argoverse}& \multirow{3}{*}{2} 
& HiVT & 4.158 & 8.368 & 0.846 & 1.085 & 1.712 & 0.249 \\
& &MOE+HiVT &  3.312 & 6.840 & 0.794 & 0.939 & 1.413 & 0.177 \\ 
                           &                    &Distill+HiVT & 3.251 & 6.638 & 0.771 & 0.968 & 1.502 & 0.185 \\ 
                           &                    &\textbf{ITPNet+HiVT} &  \textbf{2.631} & \textbf{5.703} & \textbf{0.757} & \textbf{0.819} & \textbf{1.218} & \textbf{0.141} \\
\cline{2-9}
    
                           & \multirow{3}{*}{5} 
                           & HiVT & 2.510 & 5.523 & 0.747 & 0.809 & 1.203 & 0.137 \\
                           & &MOE+HiVT &  2.562 & 5.607 & 0.776 & 0.784 & 1.221 & 0.134 \\ 
                           &                    &Distill+HiVT &  2.465 & 5.452 & 0.756 & 0.796 & 1.248 & 0.139 \\ 
                           &                    &\textbf{ITPNet+HiVT} &  \textbf{2.410} & \textbf{5.257} & \textbf{0.738} & \textbf{0.748} & \textbf{1.132} & \textbf{0.122} \\
\cline{2-9}
                           & \multirow{3}{*}{10} 
                           & HiVT & 2.441 & 5.238 & 0.741 & 0.736 & 1.121 & 0.121 \\
                           & &MOE+HiVT &  2.357 & 5.141 & 0.733 & 0.726 & 1.101 & 0.117 \\ 
                           &                    &Distill+HiVT &  2.224 & 5.039 & 0.726 & 0.731 & 1.118 & 0.119 \\ 
                           &                    &\textbf{ITPNet+HiVT} &  \textbf{2.190} & \textbf{4.792} & \textbf{0.716} & \textbf{0.718} & \textbf{1.088} & \textbf{0.113} \\
\bottomrule[1.3pt]
\end{tabular}
\label{table:dt}
\end{table*}

\noindent\textbf{Ablation Study}.
We conduct ablation studies  on the Argoverse dataset, and we employ HiVT \citep{zhou2022hivt} as the backbone. 
Table \ref{table:ablation1} shows the results.
When the $\mathcal{L}_{rec}$ is applied to the loss function, our method significantly improves the performance. This indicates the effectiveness of our proposed backward forecast mechanism for predicting the latent feature representations of previous unobserved trajectory locations. The loss $\mathcal{L}_{cts}$ further boosts the performance of the model, demonstrating the self-supervised task is meaningful.
Moreover, our method can further improve the performance when integrating our NRRFormer, underscoring the effectiveness of our NRRFormer in filtering out noise and redundancy from the predicted unobserved latent features.

\begin{figure*}[t]
    \centering
    \subfigure{
		\begin{minipage}[t]{0.18\linewidth}
			\centering
			\includegraphics[width=1\linewidth]{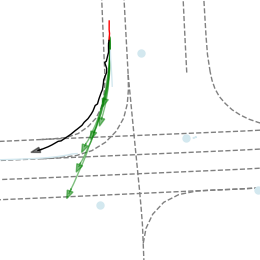}
		\end{minipage}
	}
    \subfigure{
		\begin{minipage}[t]{0.18\linewidth}
			\centering
			\includegraphics[width=1\linewidth]{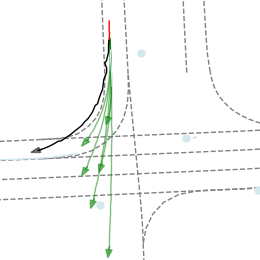}
		\end{minipage}
	}
 \subfigure{
		\begin{minipage}[t]{0.18\linewidth}
			\centering
			\includegraphics[width=1\linewidth]{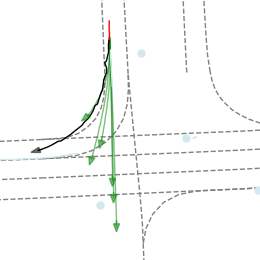}
		\end{minipage}
	}
 \subfigure{
		\begin{minipage}[t]{0.18\linewidth}
			\centering
			\includegraphics[width=1\linewidth]{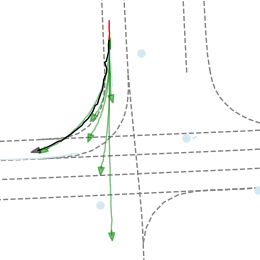}
		\end{minipage}
	}
\\
 \subfigure{
		\begin{minipage}[t]{0.18\linewidth}
			\centering
			\includegraphics[width=1\linewidth]{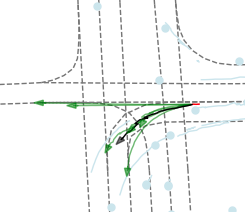}
		\end{minipage}
	}
    \subfigure{
		\begin{minipage}[t]{0.18\linewidth}
			\centering
			\includegraphics[width=1\linewidth]{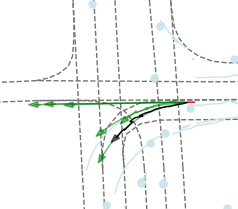}
		\end{minipage}
	}
 \subfigure{
		\begin{minipage}[t]{0.18\linewidth}
			\centering
			\includegraphics[width=1\linewidth]{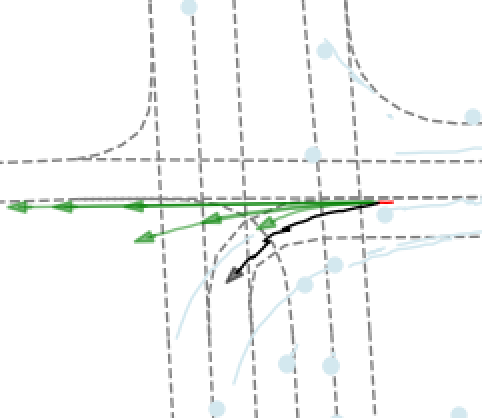}
		\end{minipage}
	}
 \subfigure{
		\begin{minipage}[t]{0.18\linewidth}
			\centering
			\includegraphics[width=1\linewidth]{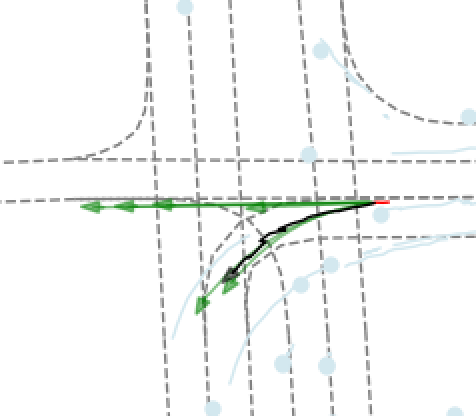}
		\end{minipage}
	}
\\
\setcounter{subfigure}{0}
 \subfigure[HiVT]{
		\begin{minipage}[t]{0.18\linewidth}
			\centering
			\includegraphics[width=1\linewidth]{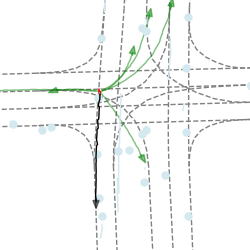}
		\end{minipage}
	}
\subfigure[MOE+HiVT]{
		\begin{minipage}[t]{0.18\linewidth}
			\centering
			\includegraphics[width=1\linewidth]{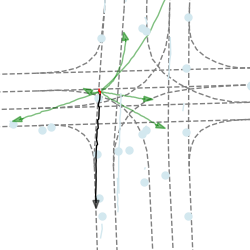}
		\end{minipage}
	}
    \subfigure[Distill+HiVT]{
		\begin{minipage}[t]{0.18\linewidth}
			\centering
			\includegraphics[width=1\linewidth]{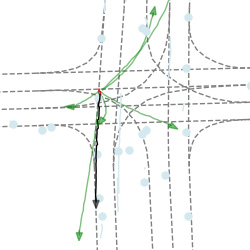}
		\end{minipage}
	}
 \subfigure[ITPNet+HiVT]{
		\begin{minipage}[t]{0.18\linewidth}
			\centering
			\includegraphics[width=1\linewidth]{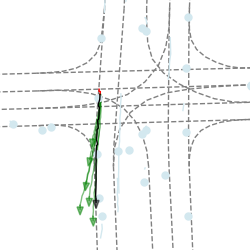}
		\end{minipage}
	}
\vspace{-0.15in}
\caption{Qualitative results of a) HiVT, b) MOE+HiVT, c) Distill+HiVT, d) ITPNet+HiVT on Argoverse. The observed historical trajectories are shown in red, the ground-truth future trajectories are shown in black, and the predicted multi-modal future trajectories are shown in green.}
	\label{fig:qualitative}
  \vspace{-0.15in}
\end{figure*}

\noindent\textbf{Analysis of Different Lengths of Unobserved Trajectories}.
We investigate the influence of different lengths $N$ of unobserved trajectory locations on instantaneous trajectory prediction.
We use HiVT as the backbone on the Argoverse dataset.
The results are listed in Table \ref{table:ablation}.
Note that when NRRFormer is not used, we directly concatenate the predicted unobserved features with observed features for future trajectory prediction.
As $N$ increases, the performance of the model is gradually improved. This reveals that predicting more latent feature representations can introduce more useful information, and thus be beneficial to trajectory prediction.
However, when $N$ exceeds a certain value ($N > 3$), the performance deteriorates. This is attributed to the introduction of noise and redundancy when predicting a longer feature sequence. When NRRFormer is enabled, the performance of our method is consistently improved as $N$ increases. This illustrates our NRRFormer module can indeed filter out redundant and noisy information, demonstrating its effectiveness.

\noindent\textbf{Results of Varied Lengths of Observed Trajectories}.
To further demonstrate the effectiveness of our method, we utilize HiVT as the backbone and conduct experiments with varied lengths of observed trajectories on Argoverse dataset. As depicted in Table \ref{table:dt}, our approach consistently exhibits superior performance across a range of $T$ when compared to the baseline methods. One interesting point is that our ITPNet+HiVT with $T=2$ achieves comparable performance to HiVT with $T=5$. This means that our method can averagely save 1.5 seconds for trajectory prediction, compared to HiVT. If a car has a driving speed of 70km/s, our method can save around 30 meters to observe the agent for trajectory prediction.

\noindent\textbf{Qualitative Results}.
We perform a visualization of the predicted multi-modal trajectories generated by MOE, Distill, HiVT, and our proposed method ITPNet+HiVT respectively on Argoverse dataset with only 2 observed locations. The results are shown in Figure \ref{fig:qualitative}.  We observe that our method exhibits diversity and more accurate trajectory prediction than other baselines in the scenario of turning and going straight. This suggests that our method can handle different driving scenarios and can achieve improved predictions with only 2 observed locations.

\section{Conclusion}
In this paper, we investigated a challenging problem of instantaneous trajectory prediction with very few observed locations. We proposed a  plug-and-play approach that backwardly predicted the latent feature representations of unobserved locations, to mitigate the issue of the lack of information. Considering the noise and redundancy in unobserved feature representations, we designed the NRRFormer to remove them and integrate the resulting filtered features and observed trajectory features into a compact query embedding for future trajectory prediction.
Extensive experimental results demonstrated that the proposed method can be effective for instantaneous trajectory prediction, and can be compatible with different trajectory prediction models.

\begin{acks}
This work was supported by the National Natural Science Foundation of China (NSFC) under Grants 62122013, U2001211.
\end{acks}

\bibliographystyle{ACM-Reference-Format}
\balance
\bibliography{ITPNet}

\newpage
\appendix

\section{Appendix}
\subsection{Implmentations of Baselines}

\noindent\textbf{MOE+HiVT and MOE+LaneGCN}. We integrate MOE \cite{sun2022human} to HiVT and LaneGCN backbones by employing the soft-pretraining with masked trajectory complement and context restoreation tasks.

\noindent\textbf{Distill+HiVT and Distill+LaneGCN.} We distill knowledge from the output of the encoder (output of Global Interaction in HiVT, and output of FusionNet in LaneGCN) and decoder (output of last hidden layer).

\begin{table*}[h]
\normalsize
\renewcommand\arraystretch{1.1}
 \setlength\tabcolsep{1.6pt}
\centering
\caption{Performance of instantaneous trajectory prediction with noised observed trajectory locations. $\sigma$ represents the standard deviation of Gaussian Noise.}
\begin{tabular}{c|c|ccc|ccc}
\toprule[1.3pt]
\multirow{2}{*}{Dataset}& \multirow{2}{*}{Methods} & \multicolumn{3}{c|}{K=1} & \multicolumn{3}{c}{K=6} \\
& & minADE & minFDE & minMR & minADE & minFDE & minMR\\ \hline

\multirow{6}{*}{Argoverse}
& MOE+HiVT & 3.312 & 6.840 & 0.794 & 0.939 & 1.413 & 0.177 \\ 
& Distill+HiVT & 3.251 & 6.638 & 0.771 & 0.968 & 1.502 & 0.185 \\ 
& ITPNet+HiVT & 2.631 & 5.703 & 0.757 & 0.819 & 1.218 & 0.141 \\ 
\cline{2-8}
& MOE+HiVT ($\sigma=0.1\text{m}$) & 3.426 & 7.114 & 0.836 & 1.002 & 1.549 & 0.202 \\ 
& Distill+HiVT ($\sigma=0.1\text{m}$) & 3.374 & 6.982 & 0.822 & 1.046 & 1.616 & 0.213 \\ 
& ITPNet+HiVT($\sigma=0.1\text{m}$) & 2.938 & 6.424 & 0.792 & 0.909 & 1.347 & 0.165 \\ 

\bottomrule[1.3pt]
\end{tabular}
\label{table:noise_comp}
\end{table*}





\subsection{Failure cases of ITPNet}
\begin{figure}[h]
    \centering
\setcounter{subfigure}{0}
\subfigure[]{
		\begin{minipage}[t]{0.46\linewidth}
			\centering
			\includegraphics[width=1\linewidth]{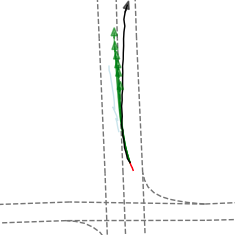}
		\end{minipage}
	}
    \subfigure[]{
		\begin{minipage}[t]{0.46\linewidth}
			\centering
			\includegraphics[width=1\linewidth]{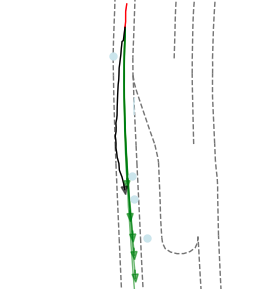}
		\end{minipage}
	}
 \\
 \subfigure[]{
		\begin{minipage}[t]{0.46\linewidth}
			\centering
			\includegraphics[width=1\linewidth]{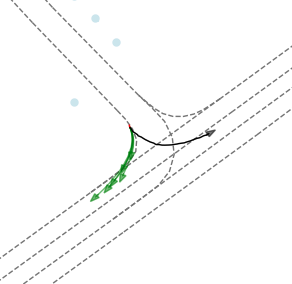}
		\end{minipage}
	}
 \subfigure[]{
		\begin{minipage}[t]{0.46\linewidth}
			\centering
			\includegraphics[width=1\linewidth]{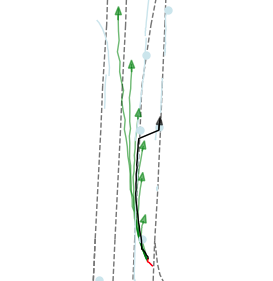}
		\end{minipage}
	}
\caption{Failure case of ITPNet+HiVT on Argoverse. The observed trajectories are shown in red, the ground-truth trajectories are shown in black, and the predicted multi-modal trajectories are shown in green.}
	\label{fig:failure}
\end{figure}

We provide failure cases of ITPNet+HiVT on Argoverse dataset, as shown in Figure \ref{fig:failure}. The model fails (1) when the future intention of the agents suddenly changes (a, d); (2) the future behavior is complex and hard to perceive from observed trajectories, such as overtaking; (3) the agent does not follow the traffic rules, such as turning left from the lane for right turns (c).

\subsection{Results with noisy observed historical trajectories}
We take into account the impact of noise in observed historical trajectories arising from perception device errors. We add Gaussian Noise $\mathcal{N}(0, \sigma)$ to observed trajectories. The experimental results, as listed in Table \ref{table:noise_comp}, reveal that our proposed ITPNet maintains its superiority over both MOE and Distill methods, even in scenarios where the observed locations exhibit noise.

\subsection{Convergence Analysis} 
\begin{figure}[h]
	\centering
	\vspace{-3mm}
	\begin{minipage}{0.46\linewidth}
		\centering
		\includegraphics[width=1\linewidth]{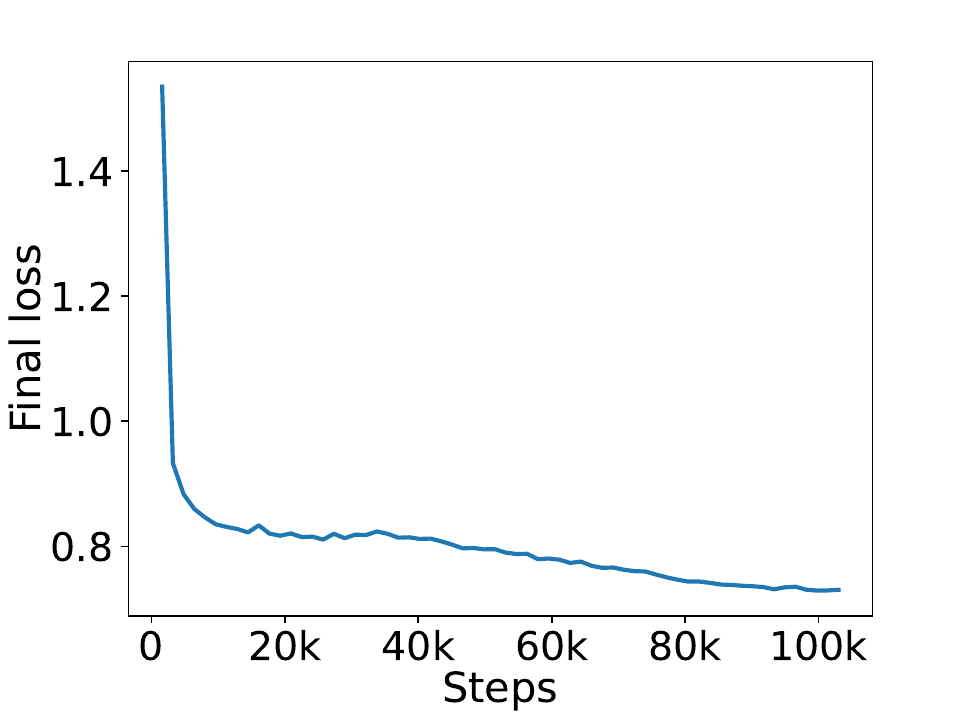}

		\label{fig:converge_hivt}
	\end{minipage}
	\begin{minipage}{0.46\linewidth}
		\centering
		\includegraphics[width=1\linewidth]{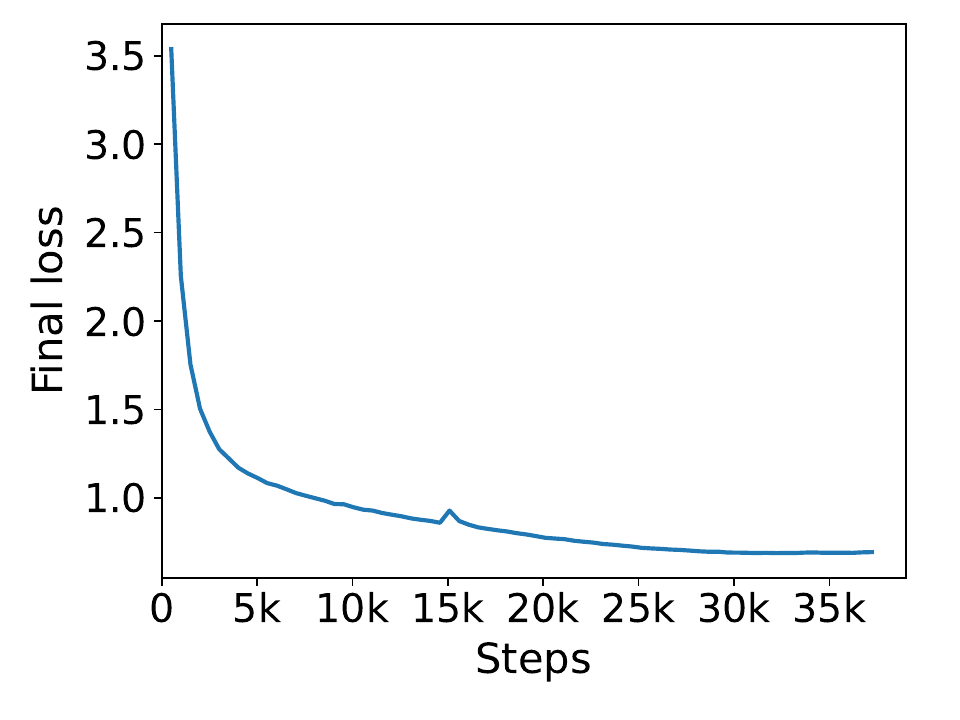}

		\label{fig:converge_nuscene}
	\end{minipage}
	\caption{Convergence analysis of our method. Left for Argoverse and right for nuScenes.}
	\label{fig:convergence}
\end{figure}

We study the convergence of our method on Argoverse and nuScenes.
The curves of the total loss of our method are shown in Figure \ref{fig:convergence}. we can see the loss decreases as the training steps, and it finally levels off.

\subsection{Details about $\mathcal{L}_{reg}$ and $\mathcal{L}_{cls}$ of backbones}
\label{app:detail_loss}
\textbf{HiVT} parameterizes the distribution of future trajectories as a mixture model where each mixture component is a Laplace distribution.
The regression loss $\mathcal{L}_{reg}$ is defined as:
\begin{equation}
\mathcal{L}_{reg} = \sum_{i=3}^{M+2} \log \frac{1}{2b}\exp(-\frac{{\vert \hat{x}_i - \mu_i \vert}}{b}),
\end{equation}
where $b$ is a learnable scale parameter of Laplace distribution, $\hat{x}^i$ is the predicted future trajectory closest to the ground-truth future trajectory and $\mu_i$ is the ground-truth future trajectory. 

The $\mathcal{L}_{cls}$ is defined as cross-entropy loss to optimize the mixing coefficients,
\begin{equation}
\mathcal{L}_{cls} = - \sum_{k=1}^K \pi^{k} \log \hat{\pi}^{k},
\end{equation}
where $\pi^k$ and $\hat{\pi}^k$ are the probability of the $k^{th}$ trajectory to be selected, and $\pi^k = 1$ if and only if $\hat{\mathbf{X}}^k $ is the predicted future trajectory closest to the ground-truth future trajectory. 

\noindent\textbf{LaneGCN} employ smooth $L_1$ loss as $\mathcal{L}_{reg}$, which is defined as,
\begin{equation}
    \mathcal{L}_{reg} = \sum_{i=3}^{M+2} \delta(\hat{x}_i - x_i),
\end{equation}
where the definition of $\delta$ is same as Equation (\ref{equ:delta}). 

The classification loss $\mathcal{L}_{cls}$ is defined as,
\begin{equation}
    \mathcal{L}_{cls} = \frac{1}{K-1} \sum_{k \neq k'} \max(0, \pi^{k} + \epsilon - \pi^{k'}),
\end{equation}
where the $k^{th}$ predicted future trajectory is the closest one to the ground-truth future trajectory. This loss pushes the closest one away from others at least by a margin $\epsilon$.


\end{document}